\documentclass{article}
\usepackage{ijcai26}

\usepackage{times}
\usepackage{soul}
\usepackage{hyperref} 
\usepackage{xurl} 
\usepackage[utf8]{inputenc}
\usepackage[small]{caption}
\usepackage{graphicx}
\usepackage{amsmath}
\usepackage{amsthm}
\usepackage{booktabs}
\usepackage{algorithm}
\usepackage{algorithmic}
\usepackage[switch]{lineno}



\title{ReusStdFlow: A Standardized Reusability Framework for Dynamic Workflow Construction in Agentic AI}

\author{
Gaoyang Zhang$^1$
\and
Shanghong Zou$^2$\thanks{The first two authors contributed equally.}\and
Yafang Wang$^1$\thanks{The corresponding author: yafang.wangcs@gmail.com} \and
He Zhang$^2$\and \\
Ruohua Xu$^{2}$\And
Feng Zhao$^1$\\
\affiliations
$^1$Accenture Information Technology, SGITG \\
$^2$Kexin Technology, CNPIEC\\
}

\newcommand{\Patk}[1]{\mbox{\Pat@$#1$}}
\newcommand{\RBPatp}[1]{\mbox{\RBP@$#1$}}

\newcommand{\NDCGatk}[1]{\mbox{\NDCG@$#1$}}
\newcommand{\ERRatk}[1]{\mbox{\ERR@$#1$}}

\newcommand{\myurl}[1]{{\url{#1}}}

\newcommand{\myparagraph}[1]{\vspace{0.2\baselineskip}\noindent{\textbf{#1}}.~}

\newcommand{\mycomment}[1]{}

\newlength{\onedigit}
\newcounter{todocount}
\begin{document}

\maketitle

\begin{abstract}
To address the ``reusability dilemma'' and structural hallucinations in enterprise Agentic AI, this paper proposes ReusStdFlow, a framework centered on a novel ``Extraction-Storage-Construction'' paradigm. The framework deconstructs heterogeneous, platform-specific Domain-Specific Languages (DSLs) into standardized, modular workflow segments. It employs a dual-knowledge architecture—integrating graph and vector databases—to facilitate synergistic retrieval of both topological structures and functional semantics. Finally, workflows are intelligently assembled using a retrieval-augmented generation (RAG) strategy. Tested on 200 real-world n8n workflows, the system achieves over 90\% accuracy in both extraction and construction. This framework provides a standardized solution for the automated reorganization and efficient reuse of enterprise digital assets.

\end{abstract}

\section{Introduction}

Driven by Large Language Model (LLM) technology, enterprise digital transformation has transitioned to an ``Agent-driven'' era~\cite{wang2025openhandsopenplatformai}, where workflow-centric logic facilitates complex task execution via structured decomposition. Consequently, workflows are now vital digital assets. However, scaling faces a ``reusability dilemma'': most workflows are scenario-specific, platform-bound (e.g., n8n~\footnote{\url{https://n8n.io/}}, Dify~\footnote{\url{https://dify.ai/}}), and lack standardized descriptions, leading to incompatibility and costly redesigns.
While existing frameworks address specific challenges—such as complex graph handling \cite{qiao2025benchmarkingagenticworkflowgeneration} or dynamic decomposition \cite{wang2025tdagmultiagentframeworkbased,gabriel2024advancingagenticsystemsdynamic}—they often suffer from linear latency or neglect reuse efficiency. For approaches primarily based on workflow construction, \cite{shen2026talmdynamictreestructuredmultiagent} cannot reuse workflows generated by other methods, and \cite{singh2025graphmemoizedreasoningfoundationsstructured} is less capable in retrieving workflow segments based on Requirements. Other modular, memory-based, or multi-agent network approaches \cite{ayala2024generatinglowcodecompleteworkflow,situmorang2025textualverifierverifytextgradstepbystep,han2025legomemmodularproceduralmemory,li2025crosstaskexperientiallearningllmbased} are hindered by structural rigidity, high computational cost, and limited compatibility, while evaluation criteria, design principles, and Human-AI co-design platforms \cite{xbench2025,asthana2025stridesystematicframeworkselecting,montazeri2025publicagentmultiagentdesignprinciples,zhang2025flowstatehumansenabling} frequently require manual intervention. Collectively, current methods prioritize forward execution over reverse decomposition, leaving a gap in autonomously reassembling static assets into adaptive workflows.

To address the ``reusability dilemma'' in enterprise Agentic AI, we propose ReusStdFlow, a standardized framework that mitigates the structural hallucinations inherent in pure generative methods by introducing a novel ``Extraction-Storage-Construction'' paradigm. Unlike existing solutions, this framework utilizes a \textbf{standardized extraction} mechanism to decompose heterogeneous, platform-specific DSLs into modular workflow segments that retain core topological logic while stripping away platform-bound redundancy. By employing a \textbf{dual-knowledge storage} strategy, these segments are persisted in a hybrid database architecture that leverages  graph database to maintain structural integrity and a vector database to enable high-performance semantic retrieval. It powers a \textbf{hybrid construction} strategy combining retrieval-augmented segment reuse with generative assembly via LLM. Empirical evaluations on 200 real-world n8n workflows demonstrate that ReusStdFlow achieves over 90\% accuracy in both extraction and construction, significantly outperforming purely generative approaches (abount 70\% accuracy) by ensuring logical closure and topological correctness through the reuse of validated subtasks.
A full demonstration of the system can be found at the following link: \url{https://note.kxsz.net/share/c8006da3-bcce-4d3a-bce3-0abbf588c20e}.

\section{Framework and Functionalities}

\begin{figure}[h]
  \centering
  \includegraphics[width=\linewidth]{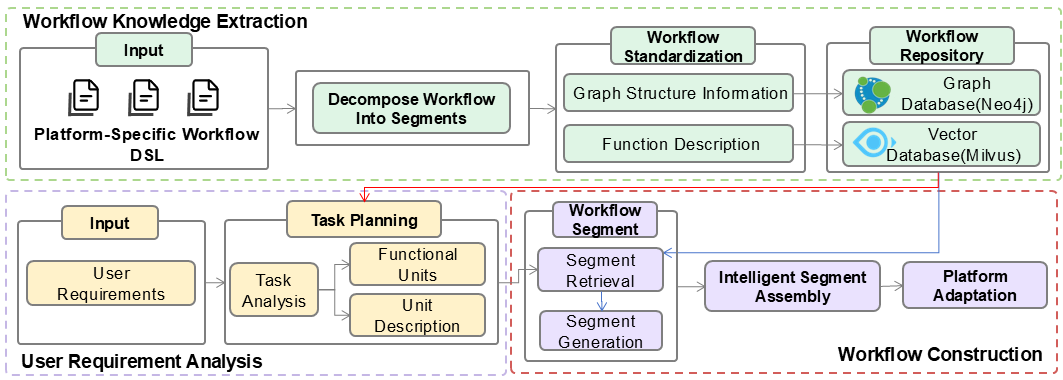}
  \caption{\small Overall architecture of the ReusStdFlow framework.}
  \label{fig:overview}
\end{figure}

\textbf{ReusStdFlow} resolves the ``reusability dilemma'' in enterprise-level Agentic AI by transforming heterogeneous legacy workflows from platforms like n8n or Dify into standardized modular assets; as shown in Figure~\ref{fig:overview}, this architecture integrates \textit{Workflow Knowledge Extraction}, \textit{User Requirement Analysis}, and \textit{Workflow Construction}.



\begin{figure}[h]
  \centering
  \includegraphics[width=\linewidth]{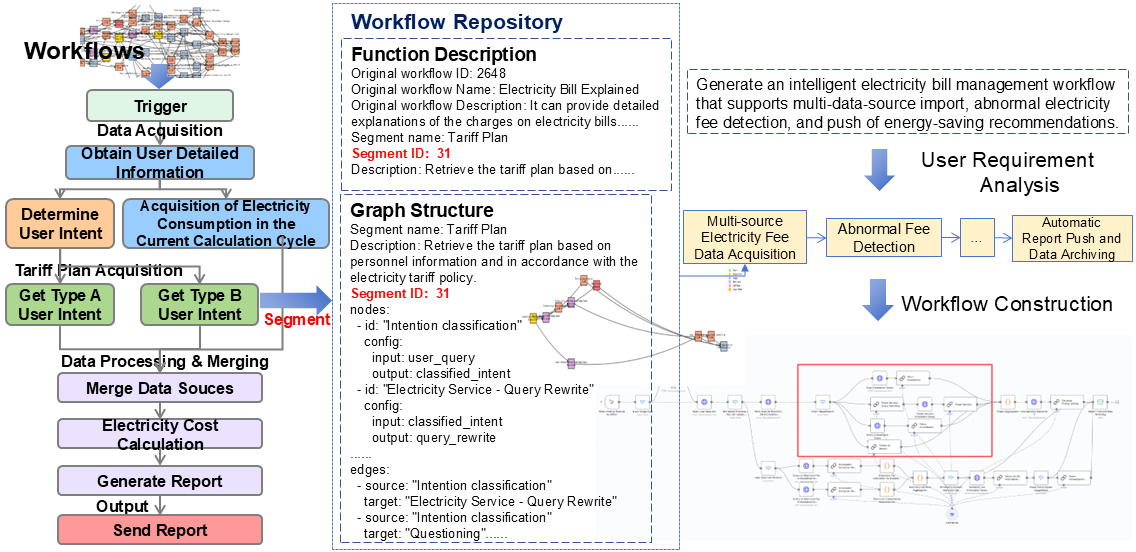}
  \caption{\small Flowchart of the entire process illustrated by a case study of electricity bill interpretation.}
  \label{fig:cases}
\end{figure}

\myparagraph{Workflow Knowledge Extraction}
This module executes the standardized parsing of platform-specific workflows into modular functional units by leveraging LLMs to deconstruct complex DSL-formatted workflows into independent segments. As illustrated in Figure~\ref{fig:cases} (left), each node represents a standardized, reusable workflow segment denoted as a directed graph $G' = (V', E')$, where $V'$ is the set of functional nodes of the workflow and $E' \subseteq V' \times V'$ defines the directed execution edges governing control and data flow. To ensure platform independence, each segment is converted into a dual-form intermediate representation consisting of a \textbf{Graph Structure}—which retains core topological relationships while stripping platform-bound redundancies like style definitions—and a \textbf{Function Description} providing a text-based semantic summary of the segment's utility. As shown in Figure~\ref{fig:cases} (middle), these components are linked via a unique \textit{segment ID}, with both representations containing metadata such as the segment ID, name, and description of the specific workflow segment. The \textit{function description} also includes the ID, name, and description of the original complete workflow. The \textit{graph structure} includes the topological information of the workflow segment and the input/output parameters of nodes.
These complementary data forms are persisted in a hybrid Dual-Knowledge architecture, utilizing a graph database for structural and topological queries and a vector database for high-performance semantic retrieval, collectively forming the\textbf{ Workflow Repository}.

\myparagraph{User Requirement Analysis}
This module bridges the gap between natural language requirements and executable tasks by utilizing the LLM to parse user input into discrete, logically coherent \textbf{functional units}. Each unit is processed by the LLM to generate a corresponding \textbf{functional unit description}, which serves as the semantic query for the subsequent retrieval phase. To enhance decomposition accuracy, the module utilizes the Workflow Repository as a reference; specifically, upon receiving a user requirement, the system retrieves the $top-k$ (where $k=10$ in our experiments) relevant complete workflows and leverages their pre-decomposed segments—generated during the knowledge extraction phase—to provide contextual guidance for the LLM in partitioning the task into logically coherent units. As demonstrated in the ``electricity bill interpretation'' scenario (see Figure~\ref{fig:cases}, right), this analytical process ensures that requirements are decomposed into hierarchically clear functional units, providing the necessary foundation for subsequent retrieval and expansion.

\myparagraph{Workflow Construction} 
As the core engine of \textit{ReusStdFlow}, this module synthesizes executable workflows via two technical routes: \textbf{retrieval-augmented assembly} and \textbf{generative assembly}. 
For each \textit{functional unit} identified in the \textit{requirement analysis} phase, the system performs semantic similarity matching within a vector database. The \textit{functional unit description} serves as the query vector to retrieve the top $k$ ($k=10$) candidate standardized segments, constrained by a similarity threshold $\theta > 0.6$. Subsequently, the corresponding graph structures are cross-referenced and retrieved from the graph database via their unique segment IDs. In cases where no candidates satisfy the threshold $\theta$, the system activates a generative route to synthesize a compliant new segment.
To ensure logical closure, the LLM executes an intelligent assembly process by analyzing parameter compatibility between adjacent segments and automatically inserting connecting nodes to bridge gaps in data dependence. Finally, the platform adaptation module augments the standardized workflow with platform-specific configurations, such as start/end nodes and canvas parameters, resulting in a complete file ready for direct deployment on platforms like n8n (see Figure~\ref{fig:cases}, bottom right).

\section{Demonstration and Applications}

\begin{figure}[h]
  \centering
  \includegraphics[width=\linewidth]{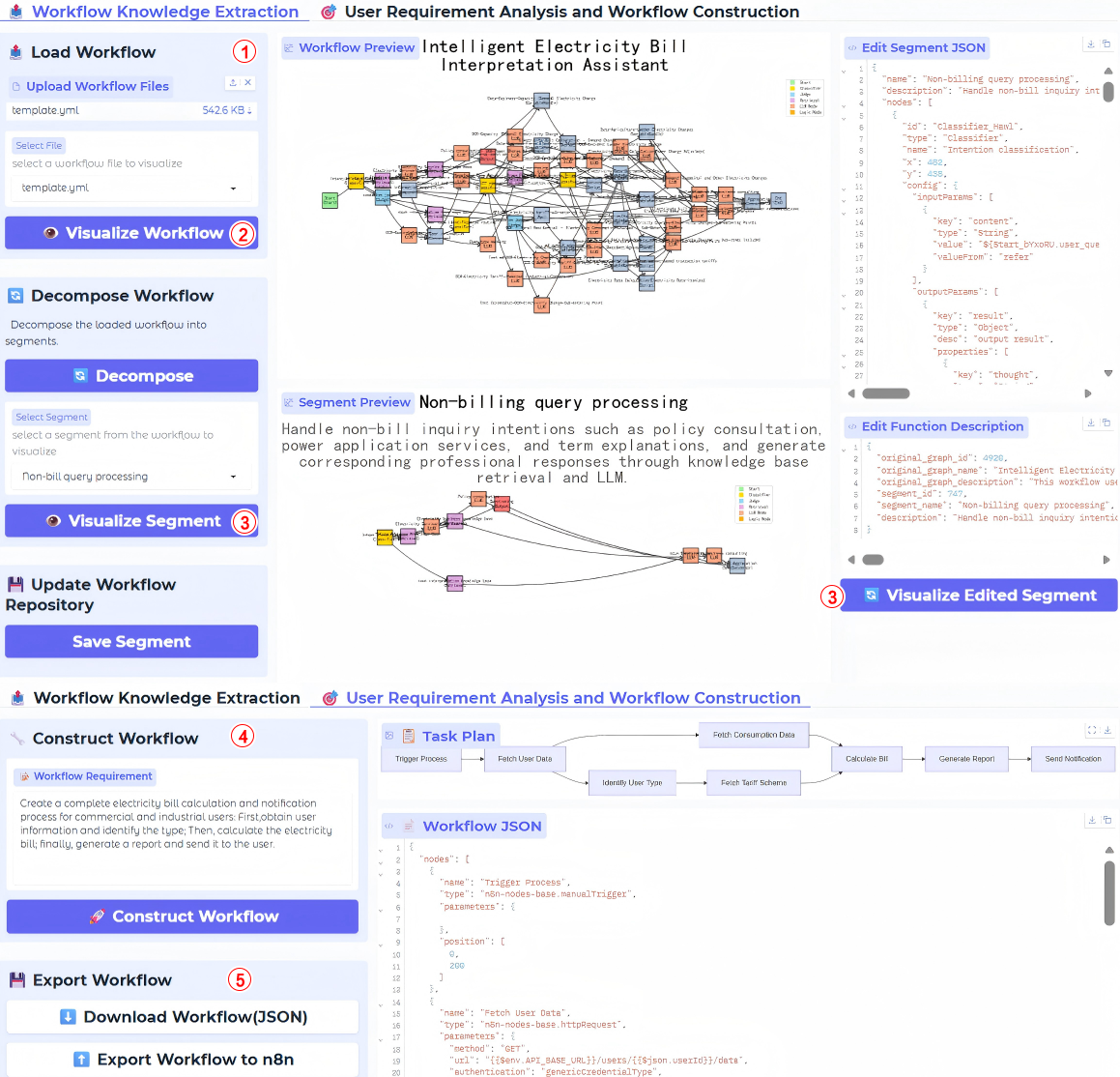}
  \caption{\small Screenshot of the ReusStdFlow interface.}
  \label{fig:screen}
\end{figure}


\myparagraph{Build Workflow Repository} As shown in Figure~\ref{fig:screen}, to build the Workflow Repository, users initiate the process by performing a batch upload of workflow files (in .yml or .yaml format) via the ``Upload Workflow Files'' button. Once uploaded, a specific workflow can be selected from the dropdown menu for full-scale visualization in the ``Workflow Preview'' panel.
By executing the ``Decompose'' command, the system automatically partitions the workflow into functional segments. Users can then inspect these segments through the ``Segment Preview'', where structural JSON data (i.e., \textit{graph structure}) and auto-generated \textit{function descriptions} are populated in their respective editors for fine-tuning. After editing, users can validate modifications via the ``Visualize Edited Segment'' button. Finalized segments are committed to the repository by clicking ``Save Segment'', ensuring a high-quality library of reusable components.
  


\myparagraph{Use Workflow Repository} Users input functional specifications into the ``Workflow Requirement'' area. Upon triggering ``Construct Workflow'', the system performs a dual-track operation:
(1) the requirement is analyzed into core functional units, displayed in the ``Task Plan'' panel, and
(2) a corresponding workflow which is synthesized and rendered in the ``Workflow JSON'' panel.
As for post-construction, the system supports seamless deployment, allowing users to either download the raw JSON configuration or directly interface with the n8n platform via the ``Export Workflow to n8n" one-click integration.


\myparagraph{Applications}
 The proposed \textit{Workflow Repository} serves as the strategic engine for an \textbf{Enterprise Knowledge Hub}, systematically mining high-value insights from legacy workflow assets to cultivate a standardized library of reusable components. In complex corporate environments, AI agents are frequently developed across fragmented departments and heterogeneous platforms; yet, these workflows often share significant underlying functional logic. By leveraging the \textit{Workflow Knowledge Extraction} module, operators can autonomously process these diverse assets, deconstructing platform-specific DSLs into standardized functional segments  to facilitate scalable repository expansion.
On the other hand, to streamline workflow development, we provide dedicated plugins for platforms like n8n that transform functional descriptions into ready-to-use drafts. By integrating pre-configured enterprise APIs and internal services, these plugins eliminate repetitive manual setups. This approach not only lowers the technical barrier for non-developers and ensures cross-departmental consistency but also maximizes the reuse of legacy assets, significantly accelerating the enterprise automation cycle.

\section{Experiments and Implementations}

 \myparagraph{Dataset} This study utilizes an open-source dataset derived from n8n~\footnote{\url{https://n8n.io/workflows/}}, comprising 200 complete workflow samples across six primary domains: Chat Workflows, Document Ops, Video Creation, API Integration, Data Processing, and Automated Workflows. Each workflow was parsed by a LLM to generate functional description. The dataset exhibits a high density of functional segments with similar logic (e.g., data validation, exception handling), indicating significant potential for reuse.

\myparagraph{Evaluation}
The manual evaluation of the \textit{workflow knowledge extraction} module yielded an accuracy rate exceeding 90\%, accounting for the validity of both nodes and edges. Error analysis revealed two primary failure modes in LLM-based parsing: (1) node omission, which prevents the reconstruction of the complete original workflow, and (2) functional misallocation, where nodes are assigned to irrelevant units. Extracted segments were archived in a dedicated Workflow Repository $R$.
To evaluate \textit{requirement analysis} and \textit{workflow construction}, the functional descriptions of the complete original workflows were utilized as user requirements, prompting the system to reconstruct workflows using segments from $R$. Manual verification yielded a construction accuracy exceeding 90\%. The primary performance bottleneck was identified within the retrieval-matching phase, characterized by occasional semantic imprecision. A comparative experiment using a zero-shot generative LLM approach (without repository segments) achieved only approximately 70\% accuracy, with failures primarily attributed to erroneous edge directions and inconsistent node relationships.

\myparagraph{System Implementation}
 The backend is developed in Python with a Gradio-based frontend~\footnote{\url{https://www.gradio.app/}}. The core logic is powered by DeepSeek-V3.2~\footnote{\url{https://www.deepseek.com/en/}}. For data management, Neo4j~\footnote{\url{https://neo4j.com/}} is employed as a graph database to support structural queries, while Milvus~\footnote{\url{https://milvus.io/}} serves as the vector database for high-performance semantic retrieval.

\section{Conclusion and Outlook}

ReusStdFlow introduces a novel ``Extraction-Storage-Construction'' paradigm to solve the reusability dilemma and structural hallucinations in enterprise Agentic AI. By deconstructing platform-specific DSLs into modular segments and leveraging a hybrid Neo4j/Milvus storage strategy, the framework enables high-fidelity, retrieval-augmented workflow synthesis. Validated on 200 real-world workflows, ReusStdFlow achieves over 90\% accuracy in maintaining topological integrity—significantly outperforming purely generative methods (around 70\%). This framework maximizes legacy asset value and drives a leap in enterprise automation efficiency. 
Future work will evolve the repository into a Standardized Skill Library~\footnote{\url{https://agentskills.io/home}}, constructing segments as independent ``Skills'' with defined semantic I/O schemas. This elevates assets from low-code snippets to modular skills supporting high-level ``Vibe'' calls, integrating ReusStdFlow with the Vibe Coding paradigm where natural language intent replaces manual configuration.
\newpage

\appendix




\bibliographystyle{named}
\bibliography{ijcai26}

@article{qiao2025benchmarkingagenticworkflowgeneration,
  author       = {Shuofei Qiao and
                  Runnan Fang and
                  Zhisong Qiu and
                  Xiaobin Wang and
                  Ningyu Zhang and
                  Yong Jiang and
                  Pengjun Xie and
                  Fei Huang and
                  Huajun Chen},
  title        = {Benchmarking Agentic Workflow Generation},
  booktitle    = {Workshop on Reasoning and Planning for Large Language Models},
  volume       = {abs/2410.07869},
  year         = {2025},
  url          = {https://openreview.net/forum?id=1VU7zLtQyW},
  doi          = {10.48550/ARXIV.2410.07869},
  eprinttype    = {arXiv},
  eprint       = {2410.07869},
  bibsource    = {dblp computer science bibliography, https://dblp.org}
}

@article{wang2025tdagmultiagentframeworkbased,
  author       = {Yaoxiang Wang and
                  Zhiyong Wu and
                  Junfeng Yao and
                  Jinsong Su},
  title        = {{TDAG:} {A} multi-agent framework based on dynamic Task Decomposition
                  and Agent Generation},
  journal      = {Neural Networks},
  volume       = {185},
  pages        = {107200},
  year         = {2025},
  url          = {https://doi.org/10.1016/j.neunet.2025.107200},
  doi          = {10.1016/J.NEUNET.2025.107200},
  bibsource    = {dblp computer science bibliography, https://dblp.org}
}

@article{gabriel2024advancingagenticsystemsdynamic,
  author       = {Adrian Garret Gabriel and
                  Alaa Alameer Ahmad and
                  Shankar Kumar Jeyakumar},
  title        = {Advancing Agentic Systems: Dynamic Task Decomposition, Tool Integration
                  and Evaluation using Novel Metrics and Dataset},
  booktitle    = {The Thirteenth International Conference on Learning Representations},
  volume       = {abs/2410.22457},
  year         = {2025},
  url          = {https://openreview.net/forum?id=OJd3ayDDoF},
  doi          = {10.48550/ARXIV.2410.22457},
  eprinttype   = {arXiv},
  eprint       = {2410.22457},
  bibsource    = {dblp computer science bibliography, https://dblp.org}
}

@inproceedings{wang2025openhandsopenplatformai,
  author       = {Xingyao Wang and
                  Boxuan Li and
                  Yufan Song and
                  Frank F. Xu and
                  Xiangru Tang and
                  Mingchen Zhuge and
                  Jiayi Pan and
                  Yueqi Song and
                  Bowen Li and
                  Jaskirat Singh and
                  Hoang H. Tran and
                  Fuqiang Li and
                  Ren Ma and
                  Mingzhang Zheng and
                  Bill Qian and
                  Yanjun Shao and
                  Niklas Muennighoff and
                  Yizhe Zhang and
                  Binyuan Hui and
                  Junyang Lin and
                  et al.},
  title        = {OpenHands: An Open Platform for {AI} Software Developers as Generalist
                  Agents},
  booktitle    = {The Thirteenth International Conference on Learning Representations},
  publisher    = {OpenReview.net},
  year         = {2025},
  url          = {https://openreview.net/forum?id=OJd3ayDDoF},
  bibsource    = {dblp computer science bibliography, https://dblp.org}
}

@misc{asthana2025stridesystematicframeworkselecting,
      title={STRIDE: A Systematic Framework for Selecting AI Modalities -- Agentic AI, AI Assistants, or LLM Calls}, 
      author={Shubhi Asthana and Bing Zhang and Chad DeLuca and Ruchi Mahindru and Hima Patel},
      year={2025},
      eprint={2512.02228},
      archivePrefix={arXiv},
      primaryClass={cs.AI},
      url={https://arxiv.org/abs/2512.02228}, 
}

@article{xbench2025,
  author       = {Kaiyuan Chen and
                  Yixin Ren and
                  Yang Liu and
                  Xiaobo Hu and
                  Haotong Tian and
                  Tianbao Xie and
                  Fangfu Liu and
                  Haoye Zhang and
                  Hongzhang Liu and
                  Yuan Gong and
                  Chen Sun and
                  Han Hou and
                  Hui Yang and
                  James Pan and
                  Jianan Lou and
                  Jiayi Mao and
                  Jizheng Liu and
                  Jinpeng Li and
                  Kangyi Liu and
                  Kenkun Liu and
                  Rui Wang and
                  Run Li and
                  Tong Niu and
                  Wenlong Zhang and
                  Wenqi Yan and
                  Xuanzheng Wang and
                  Yuchen Zhang and
                  Yi{-}Hsin Hung and
                  Yuan Jiang and
                  Zexuan Liu and
                  Zihan Yin and
                  Zijian Ma and
                  Zhiwen Mo},
  title        = {xbench: Tracking Agents Productivity Scaling with Profession-Aligned
                  Real-World Evaluations},
  journal      = {CoRR},
  volume       = {abs/2506.13651},
  year         = {2025},
  url          = {https://doi.org/10.48550/arXiv.2506.13651},
  doi          = {10.48550/ARXIV.2506.13651},
  eprinttype    = {arXiv},
  eprint       = {2506.13651},
  bibsource    = {dblp computer science bibliography, https://dblp.org}
}

@article{ayala2024generatinglowcodecompleteworkflow,
  author       = {Orlando Marquez Ayala and
                  Patrice B{\'{e}}chard},
  title        = {Generating a Low-code Complete Workflow via Task Decomposition and
                  {RAG}},
  journal      = {CoRR},
  volume       = {abs/2412.00239},
  year         = {2024},
  url          = {https://doi.org/10.48550/arXiv.2412.00239},
  doi          = {10.48550/ARXIV.2412.00239},
  eprinttype    = {arXiv},
  eprint       = {2412.00239},
  bibsource    = {dblp computer science bibliography, https://dblp.org}
}

@article{situmorang2025textualverifierverifytextgradstepbystep,
  author       = {Eugenius Mario Situmorang and
                  Adila Alfa Krisnadhi and
                  Ari Wibisono},
  title        = {TextualVerifier: Verify TextGrad Step-by-Step},
  journal      = {CoRR},
  volume       = {abs/2511.03739},
  year         = {2025},
  url          = {https://doi.org/10.48550/arXiv.2511.03739},
  doi          = {10.48550/ARXIV.2511.03739},
  eprinttype    = {arXiv},
  eprint       = {2511.03739},
  bibsource    = {dblp computer science bibliography, https://dblp.org}
}

@article{montazeri2025publicagentmultiagentdesignprinciples,
  author       = {Sina Montazeri and
                  Yunhe Feng and
                  Kewei Sha},
  title        = {PublicAgent: Multi-Agent Design Principles From an LLM-Based Open
                  Data Analysis Framework},
  journal      = {CoRR},
  volume       = {abs/2511.03023},
  year         = {2025},
  url          = {https://doi.org/10.48550/arXiv.2511.03023},
  doi          = {10.48550/ARXIV.2511.03023},
  eprinttype    = {arXiv},
  eprint       = {2511.03023},
  bibsource    = {dblp computer science bibliography, https://dblp.org}
}

@article{han2025legomemmodularproceduralmemory,
  author       = {Dongge Han and
                  Camille Couturier and
                  Daniel Madrigal D{\'{\i}}az and
                  Xuchao Zhang and
                  Victor R{\"{u}}hle and
                  Saravan Rajmohan},
  title        = {LEGOMem: Modular Procedural Memory for Multi-agent {LLM} Systems for
                  Workflow Automation},
  booktitle    = {The 25th International Conference on Autonomous Agents and Multi-Agent Systems},
  volume       = {abs/2510.04851},
  year         = {2025},
  url          = {https://openreview.net/forum?id=AGrsS2QgPC},
  doi          = {10.48550/ARXIV.2510.04851},
  eprinttype    = {arXiv},
  eprint       = {2510.04851},
  bibsource    = {dblp computer science bibliography, https://dblp.org}
}

@article{li2025crosstaskexperientiallearningllmbased,
  author       = {Yilong Li and
                  Chen Qian and
                  Yu Xia and
                  Ruijie Shi and
                  Yufan Dang and
                  Zihao Xie and
                  Ziming You and
                  Weize Chen and
                  Cheng Yang and
                  Weichuan Liu and
                  Ye Tian and
                  Xuantang Xiong and
                  Lei Han and
                  Zhiyuan Liu and
                  Maosong Sun},
  title        = {Cross-Task Experiential Learning on LLM-based Multi-Agent Collaboration},
  journal      = {CoRR},
  volume       = {abs/2505.23187},
  year         = {2025},
  url          ={https://doi.org/10.48550/arXiv.2505.23187},
  doi          = {10.48550/ARXIV.2505.23187},
  eprinttype    = {arXiv},
  eprint       = {2505.23187},
  bibsource    = {dblp computer science bibliography, https://dblp.org}
}

@article{zhang2025flowstatehumansenabling,
 author       = {Helena Zhang and
                  Jakobi Haskell and
                  Yosef Frost},
  title        = {Flow State: Humans Enabling {AI} Systems to Program Themselves},
  journal      = {CoRR},
  volume       = {abs/2504.03771},
  year         = {2025},
  url          = {https://doi.org/10.48550/arXiv.2504.03771},
  doi          = {10.48550/ARXIV.2504.03771},
  eprinttype    = {arXiv},
  eprint       = {2504.03771},
  bibsource    = {dblp computer science bibliography, https://dblp.org}
}

@article{shen2026talmdynamictreestructuredmultiagent,
  author       = {Ming{-}Tung Shen and
                  Yuh{-}Jzer Joung},
  title        = {{TALM:} Dynamic Tree-Structured Multi-Agent Framework with Long-Term
                  Memory for Scalable Code Generation},
  journal      = {CoRR},
  volume       = {abs/2510.23010},
  year         = {2025},
  url          = {https://doi.org/10.48550/arXiv.2510.23010},
  doi          = {10.48550/ARXIV.2510.23010},
  eprinttype    = {arXiv},
  eprint       = {2510.23010},
  bibsource    = {dblp computer science bibliography, https://dblp.org}
}

@article{singh2025graphmemoizedreasoningfoundationsstructured,
  author       = {Yash Raj Singh},
  title        = {Graph-Memoized Reasoning: Foundations Structured Workflow Reuse in
                  Intelligent Systems},
  journal      = {CoRR},
  volume       = {abs/2511.15715},
  year         = {2025},
  url          = {https://doi.org/10.48550/arXiv.2511.15715},
  doi          = {10.48550/ARXIV.2511.15715},
  eprinttype    = {arXiv},
  eprint       = {2511.15715},
  bibsource    = {dblp computer science bibliography, https://dblp.org}
}

\end{document}